%% file: main.tex
\documentclass[runningheads]{llncs}

\usepackage{accv}

\usepackage{accvabbrv}

\usepackage{graphicx}
\usepackage{booktabs}

\usepackage{algorithm}
\usepackage{algpseudocode}
\usepackage{multirow}
\usepackage{bm}
\usepackage[accsupp]{axessibility}  

\usepackage{hyperref}

\usepackage{orcidlink}

\begin{document}

\title{TongueReenact: Geometry-Anchored Tongue Synthesis for Face Reenactment}

\titlerunning{TongueReenact: Tongue Synthesis for Face Reenactment}

\author{MD Wahiduzzaman Khan\inst{1} \and
Mingshan Jia\inst{1} \and
Xiaolin Zhang\inst{2}\thanks{Corresponding author.} \and
En Yu\inst{1} \and
Kaska Musial-Gabrys\inst{1}}

\authorrunning{M.~W.~Khan et al.}

\institute{University of Technology Sydney, Australia\\
\email{arnobk511@gmail.com, mingshan.jia@uts.edu.au,}\\
\email{en.yu-1@uts.edu.au, musial.katarzyna@gmail.com}
\and
Shandong University of Science and Technology, China\\
\email{solli.zhang@gmail.com}}

\maketitle

\input{Sections/1_abstract}

\input{Sections/2_introduction}

\input{Sections/3_related_works}
\input{Sections/4_preliminaries}

\input{Sections/5_methodology}

\input{Sections/6_experiments}
\input{Sections/7_conclusion}

%
%
\bibliographystyle{splncs04}
\bibliography{main}

\input{Sections/8_appendix}
\end{document}

%% file: Sections/1_abstract.tex
\begin{abstract}
Modern face reenactment systems achieve impressive pose and expression transfer using geometry-driven representations. However, they largely ignore tongue dynamics, leading to anatomically inconsistent mouth interiors during speech and expressive motions. We introduce the first framework for cross-identity tongue dynamics transfer in face reenactment. We propose a foundation-model-assisted bootstrapping pipeline that produces a dedicated tongue segmentation model for in-the-wild reenactment without curated annotations. We further introduce a spatially constrained latent masked diffusion model for realistic tongue synthesis, with adaptive mask dilation for seamless mouth boundary transitions. Extensive experiments demonstrate improvements of more than two times over all baselines on every tongue-specific metric. We additionally propose a VLM-based evaluation protocol that replicates expert annotation at scale, confirming perceptual superiority across all ablation variants.
\keywords{Face Reenactment \and Tongue Synthesis \and Latent Diffusion 
\and Gaussian Splatting \and Generative Face Models}
\end{abstract}

%% file: Sections/2_introduction.tex
\section{Introduction}
Animatable head avatar synthesis and face reenactment have made substantial strides in recent years. Geometry-aware representations such as 3D Gaussian splatting \cite{kerbl20233d, dhamo2024headgas} and large-scale generative priors \cite{deng2024portrait4d, zhao2025x} now enable high-fidelity identity preservation with precise control over pose and expression. Despite this progress, a fundamental aspect of natural facial communication remains largely unaddressed, \ie, the tongue. During speech, the tongue is the primary articulator, shaping phonemes and driving visible oral dynamics. During expressive actions such as laughter, disgust, or surprise, it can dominate the visible oral region entirely. Yet nearly all existing reenactment and avatar systems either model the oral interior as a textureless void or leave it entirely unconstrained \cite{chu2024gpavatar, qian2024gaussianavatars, zhao2025x}. This neglect is not coincidental, as the tongue is a uniquely transient structure, fully concealed at rest, intermittently exposed during speech, and suddenly dominant in extreme expressions, making it fundamentally resistant to the static parametric modeling applied to the rest of the face. Addressing this overlooked challenge is the central motivation of our work.

\begin{figure*}[t]
  \centering
  \includegraphics[width=\linewidth]{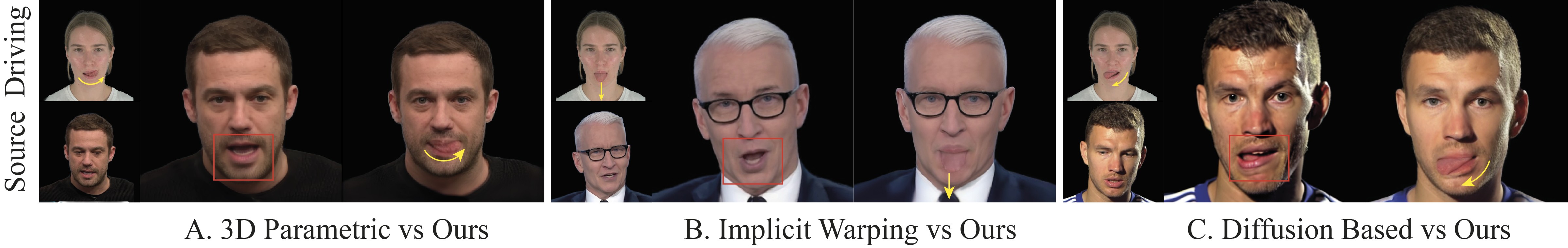}
  \caption{Existing face reenactment methods across three dominant paradigms 
  fail to transfer tongue dynamics from the driving identity to the source 
  identity. 3D parametric~\cite{chu2024gpavatar,deng2024portrait4d}, 
  implicit warping~\cite{guo2024liveportrait}, and 
  diffusion-based~\cite{zhao2025x} methods all produce tongueless mouth 
  regions (\textcolor{red}{red box}) despite a clearly visible driving tongue, whereas our 
  method successfully synthesizes accurate tongue articulation (\textcolor{yellow}{yellow arrow}) across all 
  three scenarios.}
  \label{fig:motivation}
\end{figure*}
Existing face reenactment and portrait animation methods fall broadly into three families: 3D parametric approaches, implicit warping methods, and generative animation frameworks. 3D parametric methods rely on parametric priors such as 3DMM~\cite{blanz2023morphable} or FLAME~\cite{li2017learning} to drive facial motion through explicit deformation fields, offering strong structural consistency and pose controllability. However, these parametric models are fundamentally constrained by their mesh topology, as the oral interior is an open void in the mesh with no vertices, no surface, and no geometry representing the tongue whatsoever. No amount of expression parameter tuning can synthesize a structure that does not exist in the underlying representation. Implicit warping methods~\cite{guo2024liveportrait} transfer motion by mapping source pixels to new locations using learned keypoint fields. Warping is a transformation, not a generation, and can only rearrange pixels that already exist in the source frame. If the source identity exhibits no visible tongue, no warping operation can synthesize one. Generative animation methods~\cite{zhao2025x, xie2024x} animate portraits by generating new appearance conditioned on motion from the driving video. However, these models have no explicit mechanism to identify tongue presence or transfer tongue dynamics from the driving identity. Synthesis is therefore inconsistent, either hallucinating incorrect tongue content or collapsing the mouth interior to a blurred cavity, regardless of what the driving frame exhibits. The core difficulty, shared by all three families, is not simply a lack of attention. It is the absence of any dedicated training data pairing tongue appearance with facial geometry, making the tongue an effectively invisible structure to every existing pipeline. Figure~\ref{fig:motivation} demonstrates this failure across all three paradigms.

In this paper, we present the first framework for cross-identity tongue dynamics transfer in face reenactment. Given a source frame and a driving video, our method transfers both facial motion and tongue articulation onto the source identity. The central insight of our work is that tongue synthesis cannot be treated as a generic inpainting problem. The tongue must be spatially grounded to the source face geometry, which itself changes with every driving frame. To address this, we propose a two-stage framework that explicitly decouples geometric alignment from generative tongue synthesis. In the first stage, a Gaussian splatting backbone~\cite{chu2024generalizable} conditioned on FLAME~\cite{li2017learning} parameters transfers the driving identity's pose and expression onto the source identity, producing a geometrically grounded tongue-absent render. In the second stage, a latent masked diffusion model~\cite{rombach2022high} synthesizes tongue appearance within this render, guided by dual conditioning signals from both the source geometry and the driving tongue appearance. A critical enabler of both stages is a dedicated tongue segmentation model we train from scratch via a foundation-model-assisted bootstrapping pipeline. No dedicated tongue segmentation pipeline existed for in-the-wild reenactment prior to this work. We construct it by combining SAM~\cite{kirillov2023segment} with face parsing priors to generate pseudo-labels across diverse tongue dynamics, iteratively refining and expanding the training corpus until the model generalizes reliably across identities, expressions, and partial occlusions.

The core novelty of our synthesis stage is that both the spatial constraint and the facial reference are dynamically grounded to the driving identity at every frame. The mouth region targeted for synthesis updates with the driving pose and expression, and the facial reference anchoring all non-mouth regions updates accordingly. This dynamic coupling ensures that tongue synthesis is never a free hallucination. It is geometrically constrained to preserve the source identity's appearance and the driving identity's facial structure, frame by frame. A practical challenge is that a tight segmentation mask produces visible boundary discontinuities between the synthesized tongue and the surrounding lips and teeth. We address this with adaptive latent mask dilation, which proportionally expands the active generation region based on the spatial extent of the mask, providing sufficient boundary context for seamless transitions.

Evaluating tongue synthesis poses an additional difficulty. Standard automated metrics such as FID and IoU measure distributional and geometric properties but are blind to whether the synthesized tongue looks natural, is correctly positioned within the mouth, or matches the articulation of the driving identity. Human evaluation is the gold standard for such perceptual judgments, but manual annotation at the scale required for ablation studies is prohibitively expensive. We therefore introduce a VLM-based human evaluation protocol that bridges this gap, fine-tuning a vision-language model on expert annotations to replicate human perceptual judgment at scale.

\noindent We summarize our contributions as follows:
\begin{itemize}

\item \textbf{Bootstrapped Tongue Segmentation Pipeline.} 
No dedicated tongue segmentation model existed for in-the-wild face reenactment prior to this work. We introduce a bootstrapped pipeline leveraging prompted segmentation and face parsing priors, producing a model that generalizes across identities, expressions, and partial occlusions.

\item \textbf{Geometry-Anchored Latent Masked Diffusion.} 
We introduce a spatially constrained tongue synthesis mechanism where the mouth mask and facial reference are dynamically grounded to the driving identity at every denoising step. We further propose adaptive latent mask dilation for smooth boundary transitions, eliminating discontinuities between the synthesized tongue and surrounding mouth regions.

\item \textbf{VLM-Based Tongue Quality Evaluation.} 
We propose an automated perceptual evaluation methodology for tongue synthesis, training a vision-language model on expert annotations to assess tongue quality, positioning, and naturalness at scale, filling a gap left by standard face reenactment metrics.

\end{itemize}

%% file: Sections/3_related_works.tex
\section{Related Work}

\subsection{Face Reenactment}

Face reenactment aims to transfer the head pose and facial expression of a driving subject onto a source identity while preserving appearance. Warping-based methods estimate 2D motion fields from 3DMM coefficients~\cite{blanz2023morphable} or implicit keypoints~\cite{guo2024liveportrait}, achieving scalable reenactment but failing under occluded geometry. NeRF-based approaches~\cite{deng2024portrait4d} lift reenactment into 3D but remain impractical due to slow rendering. GPAvatar~\cite{chu2024gpavatar} reconstructs animatable avatars in a single forward pass via a point-based expression field driven by 3DMM vertices. Diffusion-based methods~\cite{zhao2025x, xie2024x} demonstrate zero-shot portrait animation from implicit motion descriptors with reduced identity leakage. Despite these advances, none of the above methods model the tongue as a distinct anatomical element, leaving cross-identity tongue dynamics transfer entirely unaddressed.

\subsection{3D Gaussian Splatting for Facial Avatars}

3D Gaussian Splatting (3DGS)~\cite{kerbl20233d} provides an explicit, differentiable scene representation supporting real-time rendering at NeRF-level fidelity. Several works rig Gaussian primitives to FLAME meshes for controllable head avatars~\cite{qian2024gaussianavatars, zhou2024headstudio, dhamo2024headgas}. GAGAvatar~\cite{chu2024generalizable}, which forms the reenactment backbone of our method, generalizes this to unseen identities in a single forward pass by combining DINOv2 appearance features with FLAME-derived geometric controls. A fundamental limitation shared by all Gaussian-based methods is that primitives are conditioned on observed source pixels. If the source identity exhibits no visible tongue, no primitive can represent tongue geometry, making tongue transfer impossible within a pure Gaussian splatting framework. Our method addresses this by coupling the Gaussian stage with a geometry-anchored latent masked diffusion stage.

\subsection{Diffusion-based Face Synthesis and Inpainting}

Latent diffusion models~\cite{rombach2022high} have become the dominant paradigm for high-fidelity image synthesis. Controllable animation frameworks~\cite{zhu2024champ, hu2024animate} inject structural guidance from parametric body models via dedicated encoders. For spatially constrained synthesis, RePaint~\cite{lugmayr2022repaint} and BrushNet~\cite{ju2024brushnet} demonstrate that iterative recomposition of masked regions during the reverse diffusion process yields coherent inpainting. Our geometry-anchored diffusion extends this principle, \ie, rather than applying the mask only at initialization, we recompose the reference latent against the denoised estimate at every DDIM~\cite{song2020denoising} step, with the mask itself grounded in our driving-frame segmentation model rather than a user-supplied prior.

\subsection{Tongue and Oral Region Segmentation}

General face parsing methods such as BiSeNet~\cite{yu2018bisenet} segment facial components including lips and mouth interior, but datasets such as CelebAMask-HQ do not annotate the tongue as a distinct class. Segment Anything Model~\cite{kirillov2023segment} enables zero-shot prompted segmentation, with domain-specific adaptations such as TongueSAM~\cite{cao2023tonguesam} for clinical imagery. However, none of these generalize to the diverse lighting, extreme expressions, and cross-identity variation in reenactment datasets. We address this gap by constructing a bootstrapped annotation pipeline using SAM with face-parser-guided prompting over NERSemble~\cite{kirschstein2023nersemble}, training a lightweight BiSeNet model that operates reliably on driving frames regardless of identity or head pose.

%% file: Sections/4_preliminaries.tex
\section{Preliminaries}

\paragraph{FLAME Parametric Head Model.}
FLAME~\cite{li2017learning} is a statistical 3D head model that represents facial geometry through three low-dimensional parameter sets: shape $\boldsymbol{\beta} \in \mathbb{R}^{|\beta|}$, pose $\boldsymbol{\theta} \in \mathbb{R}^{|\theta|}$, and expression $\boldsymbol{\psi} \in \mathbb{R}^{|\psi|}$. Given these parameters, the model outputs a mesh of $N = 5{,}023$ vertices:
\begin{equation}
\mathbf{V} = \mathrm{FLAME}(\boldsymbol{\beta}, \boldsymbol{\theta}, \boldsymbol{\psi}) \in \mathbb{R}^{N \times 3}.
\end{equation}
The expression parameters $\boldsymbol{\psi}$ directly encode articulations of the jaw, lips, and surrounding musculature, providing a compact and semantically meaningful geometric prior for mouth region localization.

\paragraph{3D Gaussian Splatting.}
3D Gaussian Splatting~\cite{kerbl20233d} represents a scene as a collection of anisotropic Gaussian primitives, each defined by a center position $\boldsymbol{\mu} \in \mathbb{R}^3$, a covariance matrix $\boldsymbol{\Sigma}$, opacity $\alpha$, and spherical harmonic color coefficients. The scene is rendered by projecting the Gaussians onto the image plane and alpha-compositing them in depth order, yielding a fully differentiable rasterization pipeline that supports real-time rendering without the per-ray integration cost of NeRF-based methods.

\paragraph{Latent Diffusion Models.}
Latent diffusion models~\cite{rombach2022high} operate in the compressed latent space of a pretrained VAE. An encoder $\mathcal{E}$ maps an image $\mathbf{x}$ to a latent $\mathbf{z} = s \cdot \mathcal{E}(\mathbf{x})$, and a decoder $\mathcal{D}$ reconstructs the image as $\hat{\mathbf{x}} = \mathcal{D}(\mathbf{z})$, where $s$ is a fixed scaling factor. A denoising UNet $\epsilon_\phi$ learns to reverse a Gaussian noise process conditioned on guidance signals, and at inference the DDIM~\cite{song2020denoising} scheduler produces a deterministic trajectory from noise $\mathbf{z}_T \sim \mathcal{N}(\mathbf{0}, \mathbf{I})$ to a clean latent $\mathbf{z}_0$ in $S$ steps.

%% file: Sections/5_methodology.tex
\section{Methodology}
\begin{figure*}[t]
  \centering
  \includegraphics[width=\linewidth]{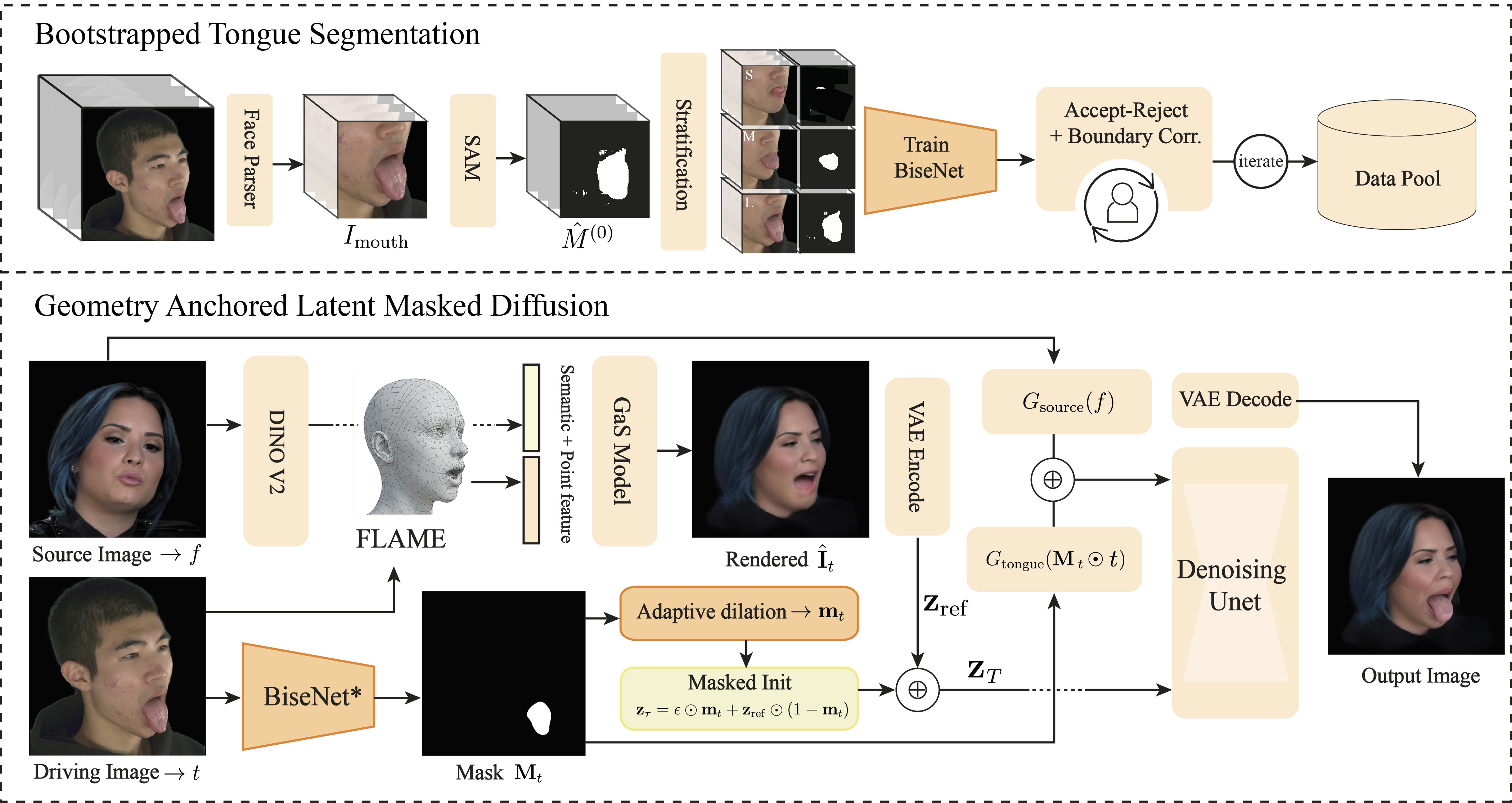}
    \caption{Overview of our pipeline. \textbf{Top:} Bootstrapped tongue segmentation training. A face parser localizes the mouth region, producing a mouth crop $I_{\text{mouth}}$. SAM generates pseudo-labels ($\hat{M}^{(0)}$), and masks are stratified into Small (S), Medium (M), and Large (L) groups. Human-in-the-loop refinement produces the training data for iterative BiSeNet* training. \textbf{Bottom:} At inference, the source image $f$ is reenacted via Gaussian splatting (GaS) conditioned on driving FLAME parameters, producing $\hat{\mathbf{I}}_t$. The driving image $t$ is segmented by BiSeNet* (bootstrapped) to yield mask $\mathbf{M}_t$, which is adaptively dilated to produce $\mathbf{m}_t$ before masked latent initialization, yielding $\mathbf{z}_T$. The reference latent $\mathbf{z}_{\mathrm{ref}}$ is encoded from $\hat{\mathbf{I}}_t$. A denoising UNet with dual guidance from $G_{\text{source}}$ and $G_{\text{tongue}}$ synthesizes the final output.
  }
  \label{fig:architecture}
\end{figure*}

Figure~\ref{fig:architecture} provides an overview of our pipeline. A foundation-model-assisted bootstrapping pipeline first produces a dedicated tongue segmentation model~(Section~\ref{sec:seg}), upon 
which the core two-stage framework is built. A Gaussian splatting backbone produces a geometry-grounded render of the source identity, which is passed to a geometry-anchored latent masked diffusion model for tongue synthesis~(Section~\ref{sec:diffusion}).

\subsection{Bootstrapped Tongue Segmentation}
\label{sec:seg}
We introduce a bootstrapped tongue segmentation pipeline that produces reliable tongue masks for in-the-wild reenactment frames without requiring manually curated annotations.

\subsubsection{Prompted Pseudo-Labeling}
For each frame $I \in \mathbb{R}^{H \times W \times 3}$ in a multi-subject video dataset, we apply a pretrained face parsing model~\cite{lee2020maskgan} to obtain a semantic label map $\mathcal{P}(I) \in \mathbb{R}^{H \times W}$. We isolate the mouth label and compute a tight axis-aligned bounding box $\mathcal{B}(I)$ to crop the mouth region $I_{\text{mouth}}$. We then run an object detector~\cite{cao2023tonguesam} to localize the tongue and obtain a bounding box prompt $\mathcal{B}_{\text{det}}$, which is passed to SAM~\cite{kirillov2023segment} to produce an initial pseudo-label:
\begin{equation} 
\hat{M}^{(0)} = \operatorname{SAM}(I_{\text{mouth}},\, \mathcal{B}_{\text{det}}).
\end{equation}
Frames with no confident detection are discarded as tongue-absent. Among the remaining frames, the candidate with the highest confidence mask is selected as the representative annotation for that timestamp.

\begin{figure*}[t]
  \centering
  \includegraphics[width=\linewidth]{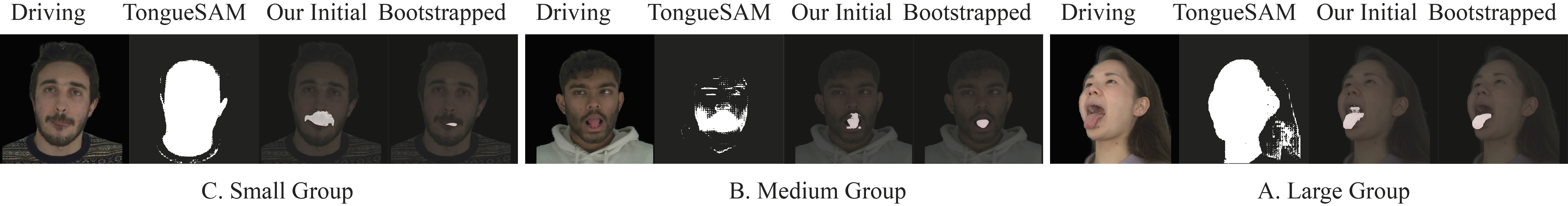}
  \caption{Tongue segmentation comparisons across three exposure groups (Small, Medium, Large). For each group we show the driving image, the TongueSAM baseline, our initial SAM-based pseudo-label, and the final bootstrapped BiSeNet* prediction. TongueSAM consistently produces erroneous full-face regions, while our bootstrapped model accurately isolates the tongue across all exposure levels.}
  \label{fig:bootstrapping}
\end{figure*}

\subsubsection{Stratified Bootstrapping}
SAM pseudo-labels $\hat{M}^{(0)}$ contain characteristic errors due to domain shift, including leakage onto lips and teeth, fragmentation under occlusion, and false positives on saturated oral tissue. To ensure balanced coverage of diverse tongue dynamics, we stratify frames by visible tongue area
\begin{equation}
A(M) = \sum_{x,y} M_{x,y},
\end{equation}
and assign each mask to one of three exposure categories:
\begin{equation}
\operatorname{Category}(M)=
\begin{cases}
\text{Small}  & \text{if } A(M) < \tau_1, \\
\text{Medium} & \text{if } \tau_1 \le A(M) < \tau_2, \\
\text{Large}  & \text{if } A(M) \ge \tau_2,
\end{cases}
\end{equation}
where $\tau_1$ and $\tau_2$ uniformly partition the observed range of mask areas. Within each category, pseudo-labels undergo two refinement stages: an accept/reject pass discarding masks with clear anatomical inconsistencies, followed by manual boundary correction for masks that are topologically correct but geometrically imprecise. The resulting refined set $\mathcal{D}^{(0)}$ forms the seed training dataset.

We train a BiSeNet~\cite{yu2018bisenet} segmentation model $\mathcal{F}_\theta$ (BiSeNet*) on $\mathcal{D}^{(0)}$. BiSeNet is chosen for its bilateral architecture: the Detail Branch preserves boundary fidelity while the Semantic Branch captures higher-level context, making it well suited to thin, irregular, and partially occluded tongue regions. We optimize using OHEM-CE loss~\cite{shrivastava2016training} on both the primary and auxiliary heads:
\begin{equation}
\mathcal{L}_{\text{seg}} 
= \mathcal{L}_{\text{OHEM}}\bigl(\mathcal{F}_\theta(I),\, M\bigr) 
+ \sum_{i=1}^{N_{\text{aux}}} 
\mathcal{L}_{\text{OHEM}}\bigl(\mathcal{F}_\theta^{(i)}(I),\, M\bigr),
\end{equation}
where $M$ is the refined pseudo-label and $N_{\text{aux}}$ is the number of auxiliary heads. The trained model $\mathcal{F}^{(0)}_\theta$ is applied to an extended set $\mathcal{V}^{\text{ext}}$ of NERSemble subjects and synthetically generated VFHQ frames, broadening identity and appearance diversity. These candidates undergo the same accept/reject and boundary correction refinement to produce $\mathcal{D}^{(1)}$, and an updated model $\mathcal{F}^{(1)}_\theta$ is trained on $\mathcal{D}^{(0)} \cup \mathcal{D}^{(1)}$. We iterate this loop until segmentation quality on held-out subjects stabilizes, measured via mask boundary consistency and inter-rater agreement (Algorithm~1, Phase~1). Qualitative results are shown in Figure~\ref{fig:bootstrapping}.

\subsection{Geometry-Anchored Latent Masked Diffusion}
\label{sec:diffusion}
The reenacted render $\hat{\mathbf{I}}_t$ provides accurate identity and pose but contains no tongue. We propose a geometry-anchored latent masked diffusion scheme. The generative process is confined exclusively to the mouth region,  as determined by the driving-frame segmentation mask. All other regions are anchored to the reenacted reference latent at every denoising step.

\subsubsection{Gaussian-Guided Reenactment}
\label{sec:reenact}
Cross-identity tongue transfer requires synthesizing tongue appearance at the precise spatial location dictated by the driving identity's facial geometry. We use a Gaussian splatting model~\cite{chu2024generalizable} to transfer the driving identity's FLAME~\cite{li2017learning} shape, pose, and expression parameters onto the source identity, producing a geometrically grounded render $\hat{\mathbf{I}}_t$. Since this render is conditioned solely on the source identity's pixels, which exhibit no visible tongue, the mouth region becomes a well-defined inpainting target for the stage that follows.

Our trained BiSeNet* (Section~\ref{sec:seg}) simultaneously operates on each driving frame $t$, producing a binary tongue segmentation mask $\mathbf{M}_t \in \{0,1\}^{H \times W}$. We extract the driving mouth crop as
\begin{equation}
\mathbf{I}_{\mathrm{mouth},t} = \mathbf{t} \odot \mathbf{M}_t,
\end{equation}
where $\odot$ denotes element-wise multiplication. Together, $\hat{\mathbf{I}}_t$ and $\mathbf{I}_{\mathrm{mouth},t}$ form a complementary conditioning signal: the former specifies where the mouth is on the source identity, and the latter specifies what the tongue looks like.

\subsubsection{Adaptive Mask Initialization}
The binary mask $\mathbf{M}_t$ from BiSeNet* is spatially expanded through an adaptive dilation procedure. Rather than a fixed-radius dilation, we compute the bounding box of the active mask region and scale the expansion proportionally to its spatial extent:
\begin{equation}
r = \max\!\left(r_{\min},\;
\Bigl\lfloor \rho \cdot \tfrac{1}{2}(h_{\mathrm{box}} + w_{\mathrm{box}}) \Bigr\rfloor\right),
\end{equation}
where $h_{\mathrm{box}}$ and $w_{\mathrm{box}}$ are the bounding box dimensions, $\rho$ is the expansion ratio, and $r_{\min}$ ensures minimal mouth coverage for frames with small tongue exposure. The dilation yields a smooth expanded mask $\tilde{\mathbf{M}}_t \in [0,1]^{H \times W}$, which is downsampled to the VAE latent resolution:
\begin{equation}
\mathbf{m}_t = \mathrm{Downsample}\!\left(
\tilde{\mathbf{M}}_t;\; \tfrac{H}{8} \times \tfrac{W}{8}
\right).
\end{equation}
Expanding in image space before downsampling avoids aliasing artifacts at latent-resolution boundaries.

The reenacted render $\hat{\mathbf{I}}_t$ is encoded by the frozen VAE encoder~\cite{rombach2022high} to obtain the reference latent:
\begin{equation}
\mathbf{z}_{\mathrm{ref}} = s \cdot \mathcal{E}(\hat{\mathbf{I}}_t),
\end{equation}
where $s = 0.18215$ is the standard VAE scaling factor. The initial latent is constructed as a region-wise composition of pure noise within the mouth region and the reference latent outside it:
\begin{equation}
\mathbf{z}_T^{(t)} = \boldsymbol{\epsilon} \odot \mathbf{m}_t
+ \mathbf{z}_{\mathrm{ref}} \odot (1 - \mathbf{m}_t),
\qquad
\boldsymbol{\epsilon} \sim \mathcal{N}(\mathbf{0}, \mathbf{I}).
\end{equation}
This confines generative uncertainty to where tongue synthesis is required, while anchoring the non-mouth region from the outset.

\subsubsection{Geometry-Anchored Denoising}
\label{sec:dual}
After each DDIM~\cite{song2020denoising} step produces an intermediate latent $\hat{\mathbf{z}}_{\tau-1}$, we recompose it against the reference:
\begin{equation}
\mathbf{z}_{\tau-1}^{(t)} = \hat{\mathbf{z}}_{\tau-1} \odot
\mathbf{m}_t + \mathbf{z}_{\mathrm{ref}} \odot (1 - \mathbf{m}_t).
\label{eq:recompose}
\end{equation}
This recomposition is applied at every step, continuously enforcing that the non-mouth region remains anchored to $\mathbf{z}_{\mathrm{ref}}$ throughout the reverse process. Since $\mathbf{m}_t$ is derived from the driving identity's geometry, the spatial constraint is geometrically precise rather than heuristic, which is the key distinction from generic latent inpainting.

The denoising UNet is conditioned on two complementary signals. The reenacted render $\hat{\mathbf{I}}_t$ is processed by a guidance encoder~\cite{zhu2024champ} to produce structural features encoding the source identity's head pose and facial geometry. The driving mouth crop $\mathbf{I}_{\mathrm{mouth},t}$ is processed by a second guidance encoder to produce appearance features encoding the driving identity's tongue texture and articulation. The two feature maps are summed and injected into the UNet at each spatial resolution level. Additionally, CLIP~\cite{radford2021learning} image embeddings from $\hat{\mathbf{I}}_t$ provide global identity conditioning through the image prompt mechanism of the diffusion backbone~\cite{rombach2022high}. The complete inference procedure is summarized in Algorithm~1 (Phase~2).

\algnewcommand\Phase{\item[\textbf{Phase}]}

\begin{figure}[t]

\footnotesize
\noindent\rule{\linewidth}{0.4pt}
\noindent\parbox{\linewidth}{\raggedright\textbf{Algorithm 1:} Cross-Identity Tongue Dynamics Transfer}
\phantomcaption
\noindent\rule{\linewidth}{0.4pt}
\begin{minipage}[t]{0.48\linewidth}
\raggedright
\begin{algorithmic}[1]
\Require $\mathcal{V}^{\text{NER}}$, $\mathcal{V}^{\text{ext}}$, $\mathcal{P}$, SAM
\Ensure BiSeNet* 
\Phase \textbf{1:} Bootstrapped Segmentation Training
\For{each $I \in \mathcal{V}^{\text{NER}}$}
    \State $I_{\mathrm{mouth}} \leftarrow \mathcal{P}(I)$
    \State $\hat{M}^{(0)} \leftarrow \operatorname{SAM}(I_{\mathrm{mouth}}, \mathcal{B}_{\text{det}})$
    \State Stratify $A(M)$ $\rightarrow$ S / M / L
\EndFor
\State Refine $\rightarrow \mathcal{D}^{(0)}$; train $\mathcal{F}^{(0)}_\theta$
\State $k \leftarrow 1$
\Repeat
    \State $\mathcal{F}^{(k-1)}_\theta$ on $\mathcal{V}^{\text{ext}}$ $\rightarrow$ refine $\mathcal{D}^{(k)}$
    \State $\mathcal{V}^{\text{ext}} \mathrel{+}= \text{GaS}(f, \text{FLAME}(t))$
    \State Train $\mathcal{F}^{(k)}_\theta$ on $\mathcal{D}^{(k-1)} \cup \mathcal{D}^{(k)}$
    \State $k \mathrel{+}= 1$
\Until{quality stabilizes}
\State BiSeNet* $\leftarrow \mathcal{F}^{(k-1)}_\theta$
\end{algorithmic}
\end{minipage}
\hfill\vline\hfill
\begin{minipage}[t]{0.48\linewidth}
\raggedright
\begin{algorithmic}[1]
\Require $f$, $\{t\}$, BiSeNet*, $\epsilon_\phi$
\Ensure $\{O_t\}$
\Phase \textbf{2:} Tongue Synthesis
\For{each driving frame $t$}
    \State $\hat{\mathbf{I}}_t \leftarrow \text{GaS}(f, \text{FLAME}(t))$
    \State $\mathbf{M}_t \leftarrow \text{BiSeNet*}(t)$
    \State $\mathbf{m}_t \leftarrow \text{AdDilate}(\mathbf{M}_t)$
    \State $\mathbf{z}_{\mathrm{ref}} \leftarrow s\cdot\mathcal{E}(\hat{\mathbf{I}}_t)$
    \State $\mathbf{z}_T \leftarrow \boldsymbol{\epsilon} \odot \mathbf{m}_t + \mathbf{z}_{\mathrm{ref}} \odot (1-\mathbf{m}_t)$
    \State $\mathbf{c} \leftarrow G^{\text{src}}(f) + G^{\text{tng}}(\mathbf{I}_{\mathrm{mouth},t})$
    \For{$\tau = T, \ldots, 1$}
        \State $\hat{\mathbf{z}}_{\tau-1} \leftarrow \epsilon_\phi(\mathbf{z}_\tau, \tau, \mathbf{c})$
        \State $\mathbf{z}_{\tau-1} \leftarrow \hat{\mathbf{z}}_{\tau-1} \odot \mathbf{m}_t$
        \Statex \hspace{6em} $+ \mathbf{z}_{\mathrm{ref}} \odot (1-\mathbf{m}_t)$
    \EndFor
    \State $O_t \leftarrow \text{VAE-Dec}(\mathbf{z}_0)$
    
\EndFor
\State \Return $\{O_t\}$;
\end{algorithmic}
\end{minipage}
\vspace{2pt}
\noindent\rule{\linewidth}{0.4pt}
\end{figure}

%% file: Sections/6_experiments.tex
\section{Experiments}

\subsection{Implementation Details}
We train BiSeNet~\cite{yu2018bisenet} with two output classes (background and tongue) for 50{,}000 iterations using SGD with momentum, learning rate $10^{-3}$, weight decay $5 \times 10^{-4}$, and linear warmup over 1{,}000 iterations. Input resolution is $512 \times 512$ with batch size 4, multi-scale augmentation in $[0.75, 2.0]$, and FP16 precision. Training data consists of NERSemble and synthetically generated VFHQ frames from our bootstrapped pipeline. For reenactment, we use the pretrained Gaussian splatting backbone~\cite{chu2024generalizable} without modification. The diffusion model is initialized from the Stable Diffusion image variations checkpoint~\cite{rombach2022high}, with the VAE and CLIP encoder frozen. Guidance encoders and the denoising UNet are trained on NERSemble for 85{,}000 iterations, batch size 8 at $512 \times 512$, using Adam ($lr{=}10^{-5}$, $\beta_1{=}0.9$, $\beta_2{=}0.999$, weight decay $10^{-2}$), noise offset 0.05, SNR $\gamma{=}5.0$, and zero-SNR training~\cite{lin2024common} with a scaled linear $\beta$ schedule ($T{=}1{,}000$ steps). Guidance encoders are trained from scratch with 320 output embedding channels. At inference we use 20 DDIM~\cite{song2020denoising} steps, $\rho{=}0.3$, and $r_{\min}{=}5$ pixels. All experiments are run on a single NVIDIA A40 (48\,GB).

\begin{figure*}[!t]
  \centering
  \includegraphics[width=\linewidth]{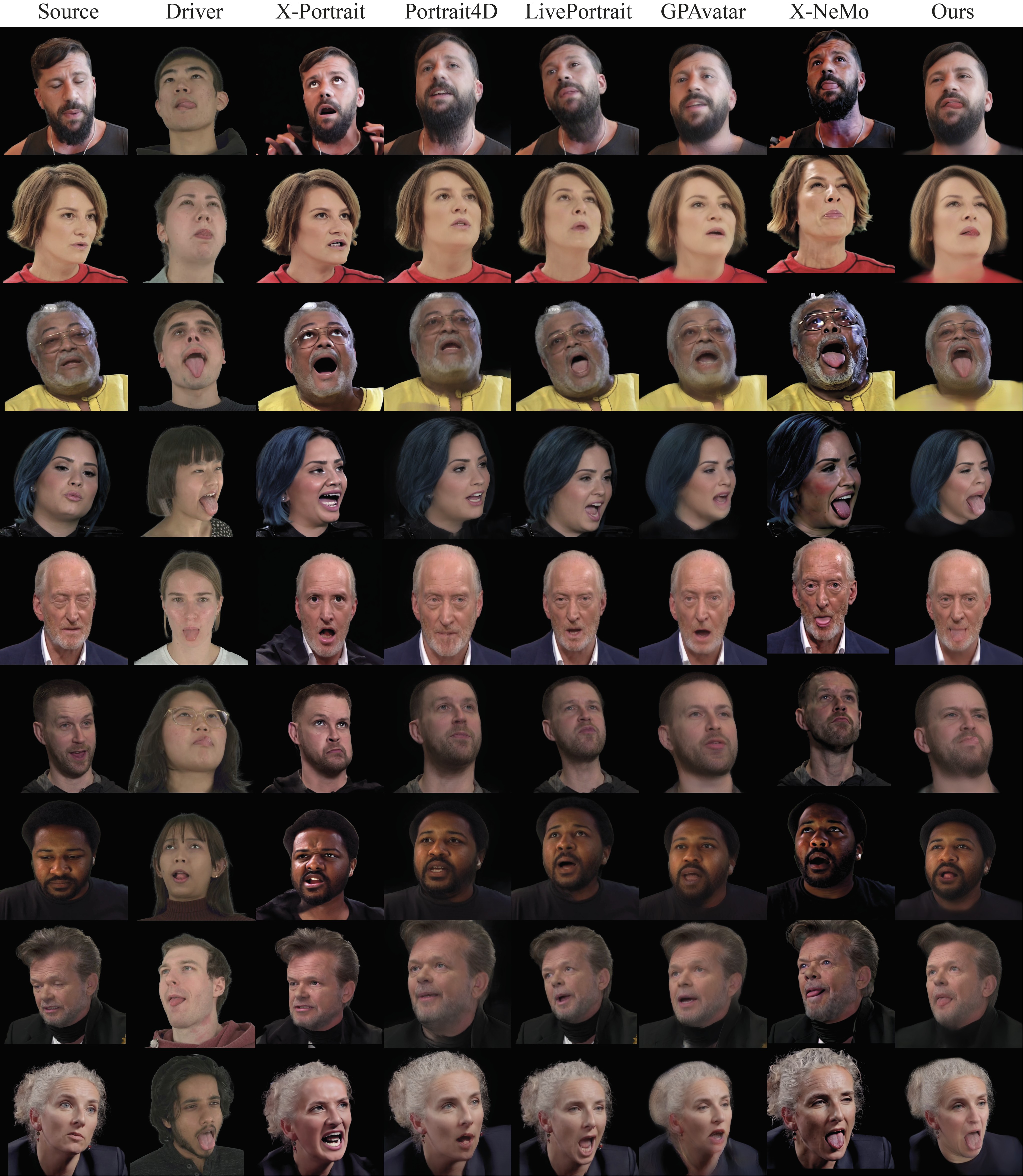}
  \caption{Qualitative comparison of cross-identity face reenactment across 
  nine source--driving pairs. Our method consistently synthesizes visible and 
  anatomically plausible tongue dynamics transferred from the driving identity, 
  whereas X-Portrait~\cite{xie2024x}, Portrait4D~\cite{deng2024portrait4d}, 
  LivePortrait~\cite{guo2024liveportrait}, GPAvatar~\cite{chu2024gpavatar}, and 
  X-NeMo~\cite{zhao2025x} fail to reproduce tongue articulation despite the 
  driving frame exhibiting a clearly visible tongue.}
  \label{fig:comparison}
\end{figure*}

\subsection{Evaluation Setup}
We evaluate on 9{,}000 cross-identity source-driving pairs from VFHQ~\cite{xie2022vfhq}, comparing against five state-of-the-art face reenactment methods: GPAvatar~\cite{chu2024gpavatar}, Portrait4D~\cite{deng2024portrait4d}, LivePortrait~\cite{guo2024liveportrait}, X-NeMo~\cite{zhao2025x}, and X-Portrait~\cite{xie2024x}, each run using their official implementations. We evaluate tongue transfer 
fidelity using four dedicated metrics: LPIPS in the tongue region~\cite{zhang2018unreasonable}, Tongue Presence (fraction of frames 
containing a detectable tongue), Tongue IoU (normalized intersection-over-union between generated and driving tongue masks), and Tongue Area Similarity (ratio of generated to driving tongue mask area). Arrows indicate the direction of improvement.

\subsection{Comparison with State-of-the-Art}
Table~\ref{tab:comparison} reports quantitative results against all baselines. Our method achieves the best performance across all four tongue transfer metrics, with a Tongue IoU of 0.3582 and Tongue Area Similarity of 0.4338, more than double the nearest competitor, confirming substantially more accurate tongue dynamics transfer. Tongue Presence of 0.7891 further demonstrates that our method reliably synthesizes visible tongue articulation where existing methods frequently produce none.

\begin{table}[t]
\centering
\caption{Comparative quantitative results. Best results are shown in \textbf{bold}.
Arrows indicate the desired direction of each metric.}
\label{tab:comparison}
\resizebox{\linewidth}{!}{%
\begin{tabular}{lrrrrrr}
\toprule
Metric & Ours & X-NeMo & GPAvatar & LivePortrait & Portrait4D & X-Portrait \\
\midrule
LPIPS (tongue) $\downarrow$    & \textbf{0.2306} & 0.3937 & 0.2890 & 0.3454 & 0.2591 & 0.4209 \\
Tongue Presence $\uparrow$            & \textbf{0.7891} & 0.7367 & 0.4274 & 0.5986 & 0.5102 & 0.5067 \\
Tongue IoU $\uparrow$                 & \textbf{0.3582} & 0.1400 & 0.1387 & 0.1603 & 0.1918 & 0.1508 \\
Tongue Area Similarity $\uparrow$     & \textbf{0.4338} & 0.1180 & 0.0713 & 0.1255 & 0.0818 & 0.1330 \\
\bottomrule
\end{tabular}}
\end{table}

\subsection{Ablation Study}
Table~\ref{tab:ablation} analyzes the contribution of each component. We evaluate three variants: removing adaptive mask dilation (w/o Dilation), removing the reenactment stage (w/o Reenactment), and removing the driving mouth crop conditioning (w/o Tongue).

Removing adaptive dilation causes a notable drop in tongue transfer metrics (Tongue IoU from 0.3956 to 0.2619, Tongue Area Similarity from 0.4779 to 0.1890), confirming that proportional mask expansion is necessary for sufficient generative freedom around the mouth boundary. Removing the reenactment stage causes tongue transfer to collapse (Tongue IoU 0.2355), demonstrating that the geometry-grounded render is essential for spatially anchoring the synthesis region. Removing tongue conditioning causes the most severe degradation across all metrics, confirming that $\mathbf{I}_{\mathrm{mouth},t}$ is the primary source of tongue appearance information.

\paragraph{Identity Preservation.}
Our method faithfully preserves identity across all non-synthesized facial regions, as confirmed by masking the mouth region prior to ArcFace~\cite{deng2019arcface} feature extraction, which reduces the identity gap between the full model and w/o Tongue variant to just 0.0007 (0.4840 vs.\ 0.4833). The small unmasked gap of 0.0239 (0.5293 vs.\ 0.5532) is entirely localized to the synthesized mouth area, an expected consequence of generating content that is entirely absent in the source frame.

\begin{table}[t]
\centering
\caption{Ablation study analyzing the contribution of each component.
Best results are shown in \textbf{bold}.}
\label{tab:ablation}
\begin{tabular}{lrrrr}
\toprule
Metric & Full (Ours) & w/o Dilation & w/o Reenactment & w/o Tongue \\
\midrule
LPIPS (tongue) $\downarrow$    & \textbf{0.2089} & 0.2793 & 0.2604 & 0.2884 \\
Tongue Presence $\uparrow$            & \textbf{0.8851} & 0.5776 & 0.7315 & 0.4397 \\
Tongue IoU $\uparrow$                 & \textbf{0.3956} & 0.2619 & 0.2355 & 0.1216 \\
Tongue Area Similarity $\uparrow$     & \textbf{0.4779} & 0.1890 & 0.2694 & 0.0631 \\
\bottomrule
\end{tabular}
\end{table}

\subsection{Temporal Consistency}

A key concern with diffusion-based synthesis is temporal flickering, as frame-by-frame generation can produce inconsistent motion across a sequence. We verify that our geometry-anchored denoising suppresses this by comparing temporal consistency between our full model and a variant without tongue synthesis over generated video sequences. Table~\ref{tab:temporal} reports T-LPIPS (perceptual similarity between consecutive frames), flow jitter (variance of inter-frame flow magnitude), and flow magnitude, all evaluated on the mouth region. Our full model achieves substantially lower flow jitter (0.218543 vs.\ 0.622901) and flow magnitude (0.513165 vs.\ 1.254073), confirming that geometry-anchored denoising produces smooth and stable tongue motion. Lower T-LPIPS further confirms that tongue synthesis does not compromise temporal coherence.

\begin{table}[t]
\centering
\caption{Temporal consistency metrics. Lower values indicate improved
smoothness and stability.}
\label{tab:temporal}
\begin{tabular}{lrr}
\toprule
Metric & With Tongue & Without Tongue \\
\midrule
T-LPIPS $\downarrow$     & \textbf{0.0149} & 0.0291 \\
Flow Jitter $\downarrow$ & \textbf{0.2185} & 0.6229 \\
Flow Magnitude   $\downarrow$  & \textbf{0.5132} & 1.2541 \\
\bottomrule
\end{tabular}
\end{table}

\subsection{VLM-Based Perceptual Evaluation}
Standard automated metrics do not capture perceptual tongue quality as judged by a human observer. We therefore fine-tune Qwen3-VL~\cite{yang2025qwen3} on 100 expert-annotated frames to replicate human judgement across four criteria: tongue quality (0--5 scale), tongue positioning (binary), best tongue, and best overall reenactment. The fine-tuned model is applied to the remaining 1{,}000 frames, yielding 1{,}100 evaluated frames in total. Full annotation details and fine-tuning procedure are provided in the supplementary material. As shown in Table~\ref{tab:human_eval}, the full model achieves a mean tongue quality rating of 3.07, is selected as best tongue in 99.9\% of frames and best overall in 100\% of frames, with correct tongue positioning in 100\% of cases. All ablation variants collapse near zero across all criteria, confirming that tongue conditioning, adaptive dilation, and geometry-anchored reenactment are each individually necessary for perceptually acceptable tongue synthesis.

\begin{table}[t]
\centering
\caption{VLM-based perceptual evaluation. \textbf{Bold} indicates best result.}
\label{tab:human_eval}
\begin{tabular}{lcccc}
\toprule
Method & Avg. Rating & Best Tongue (\%) & Best Overall (\%) & Positioning (\%) \\
\midrule
Full Model   & \textbf{3.07} & \textbf{99.9} & \textbf{100.0} & \textbf{100.0} \\
w/o Tongue   & 0.73 & 0.0 & 0.0 & 1.9 \\
w/o Dilation & 1.36 & 0.1 & 0.0 & 41.0 \\
w/o Reenact  & 1.09 & 0.0 & 0.0 & 1.6 \\
\bottomrule
\end{tabular}
\end{table}

%% file: Sections/7_conclusion.tex
\section{Conclusion}
We present a framework for cross-identity tongue dynamics transfer, a capability absent from all existing face reenactment methods. Our 
bootstrapped tongue segmentation pipeline produces reliable tongue masks without manually curated annotations. These masks drive a geometry-anchored latent masked diffusion model that synthesizes natural tongue dynamics from one identity onto another. The diffusion stage is designed to couple with any geometry-faithful reenactment backbone, making the approach broadly extensible beyond the specific model used here. Experiments on VFHQ demonstrate that our method outperforms all baselines across every tongue-specific metric, while preserving identity faithfully in all non-synthesized facial regions.

\paragraph{Limitations.}
As the diffusion stage is designed to generalize across reenactment backbones, tongue synthesis quality may vary slightly with the choice of reenactment backbone. Identity preservation in the synthesized mouth area remains slightly reduced, an inherent trade-off of generative synthesis over content entirely absent in the source frame.

%% file: Sections/8_appendix.tex
\clearpage
\setcounter{section}{0}
\renewcommand{\thesection}{\arabic{section}}

\begin{center}
{\Large \textbf{Supplementary Material for TongueReenact:\\
Geometry-Anchored Tongue Synthesis for Face Reenactment}}
\end{center}
\vspace{10pt}

\section{Extended Tongue Segmentation Evaluation}
\label{supp:seg}

We report comprehensive segmentation metrics comparing 
TongueSAM~\cite{cao2023tonguesam}, our initial seed model, 
and our final bootstrapped BiSeNet* across 6{,}344 evaluation 
frames from NERSemble~\cite{kirschstein2023nersemble}. 
Tables~\ref{tab:supp_overall} and~\ref{tab:supp_group} 
report performance on tongue-present frames. 
Table~\ref{tab:supp_absent} reports hallucination behavior 
on tongue-absent frames.

TongueSAM achieves near-zero precision, producing predicted 
masks 10--115$\times$ larger than ground truth depending on 
the exposure group. While TongueSAM achieves the highest 
recall, this is a consequence of extreme over-segmentation 
rather than accurate detection, as confirmed by area ratios 
of 10--115$\times$ ground truth. Our bootstrapped BiSeNet* 
trades recall for substantially higher precision and IoU, 
yielding masks that are geometrically accurate rather than 
over-inclusive. Our initial model substantially improves 
localization across all groups but overestimates mask area 
in the Medium group (ratio 3.07), reflecting domain shift on 
partially visible tongue frames. The final bootstrapped 
BiSeNet* achieves the best IoU and precision across all 
groups, with area ratios closest to 1.0, and suppresses 
hallucination on tongue-absent frames to 18\%, compared to 
100\% for both TongueSAM and our initial model.

\begin{table}[h]
\centering
\caption{Overall segmentation metrics on tongue-present frames 
($N = 3{,}131$).}
\label{tab:supp_overall}
\setlength{\tabcolsep}{12pt}
\begin{tabular}{lrrr}
\toprule
Metric & TongueSAM & Our Initial & Bootstrapped \\
\midrule
IoU $\uparrow$        & 0.104 & 0.597 & \textbf{0.640} \\
Precision $\uparrow$  & 0.105 & 0.653 & \textbf{0.783} \\
Recall $\uparrow$     & \textbf{0.895} & 0.872 & 0.727 \\
F1 $\uparrow$         & 0.156 & 0.711 & \textbf{0.730} \\
Area Ratio $\to 1$    & 54.46 & 1.842 & \textbf{0.979} \\
\bottomrule
\end{tabular}
\end{table}

\begin{table}[t]
\centering
\caption{Per-group segmentation metrics on tongue-present frames.}
\label{tab:supp_group}
\resizebox{\linewidth}{!}{%
\setlength{\tabcolsep}{12pt}
\begin{tabular}{llrrr}
\toprule
Group & Metric & TongueSAM & Our Initial & Bootstrapped \\
\midrule
\multirow{5}{*}{Small ($N=959$)}
    & IoU $\uparrow$       & 0.103 & 0.724 & \textbf{0.752} \\
    & Precision $\uparrow$ & 0.104 & 0.784 & \textbf{0.866} \\
    & Recall $\uparrow$    & 0.889 & \textbf{0.902} & 0.843 \\
    & F1 $\uparrow$        & 0.159 & 0.828 & \textbf{0.842} \\
    & Area Ratio $\to 1$   & 28.42 & 1.216 & \textbf{1.015} \\
\midrule
\multirow{5}{*}{Medium ($N=1{,}148$)}
    & IoU $\uparrow$       & 0.010 & 0.345 & \textbf{0.376} \\
    & Precision $\uparrow$ & 0.010 & 0.367 & \textbf{0.599} \\
    & Recall $\uparrow$    & \textbf{0.855} & 0.848 & 0.474 \\
    & F1 $\uparrow$        & 0.021 & \textbf{0.488} & 0.484 \\
    & Area Ratio $\to 1$   & 115.15 & 3.068 & \textbf{0.899} \\
\midrule
\multirow{5}{*}{Large ($N=1{,}024$)}
    & IoU $\uparrow$       & 0.210 & 0.760 & \textbf{0.832} \\
    & Precision $\uparrow$ & 0.211 & 0.852 & \textbf{0.911} \\
    & Recall $\uparrow$    & \textbf{0.945} & 0.870 & 0.902 \\
    & F1 $\uparrow$        & 0.304 & 0.851 & \textbf{0.901} \\
    & Area Ratio $\to 1$   & 10.81 & 1.052 & \textbf{1.034} \\
\bottomrule
\end{tabular}}
\end{table}

\begin{table}[t]
\centering
\caption{Mean predicted mask area (pixels) on tongue-absent 
frames. Lower values indicate fewer false positives. 
Group-level counts are small ($N=12$) and should be 
interpreted accordingly.}
\label{tab:supp_absent}
\setlength{\tabcolsep}{12pt}
\begin{tabular}{lrrr}
\toprule
Group & TongueSAM & Our Initial & Bootstrapped \\
\midrule
Medium ($N=12$)       & 35{,}470 & 1{,}500 & \textbf{0.42} \\
Large  ($N=12$)       & 65{,}606 & 2{,}241 & \textbf{85.2} \\
Overall ($N=3{,}213$) & 56{,}189 & 1{,}577 & \textbf{343} \\
\bottomrule
\end{tabular}
\end{table}

\section{VLM-Based Evaluation Details}
\label{supp:vlm}

\subsection{Annotation Protocol}

Expert annotators evaluated 100 frames across four ablation 
variants: Full Model, w/o Tongue, w/o Dilation, and w/o 
Reenactment. For each frame, annotators were presented with 
the source image, the driving image, and the four generated 
outputs side by side. Annotations covered four criteria:

\begin{itemize}
    \item \textbf{Tongue Quality.} Rated on a 0--5 scale, 
    where 0 indicates no visible tongue.
    \item \textbf{Tongue Positioning.} A binary judgement of 
    whether the tongue is correctly positioned relative to 
    the driving frame.
    \item \textbf{Best Tongue.} Which variant produces the 
    best tongue result.
    \item \textbf{Best Overall.} Which variant produces the 
    best overall reenactment.
\end{itemize}

\subsection{Fine-Tuning Setup}

We fine-tune Qwen3-VL-8B-Instruct~\cite{yang2025qwen3} on 
the 100 annotated frames using LoRA~\cite{hu2022lora}. The 
annotated set is split 80/20 into training and 
validation. Each training sample consists of 
six images: the source identity, the driving frame, and the 
four generated outputs, paired with the structured annotation 
as the target response.

LoRA is applied with rank $r=64$, scaling factor 
$\alpha=128$, and dropout 0.05. The multimodal LLM head is 
trained jointly with the LoRA adapters. Training runs for 10 
epochs with an effective batch size of 16 (per-device batch 
size 2, gradient accumulation 8), learning rate $2 \times 
10^{-4}$ with a cosine schedule and warmup ratio 0.03, weight 
decay 0.01, and maximum sequence length 2{,}048 tokens, using 
bf16 precision on a single NVIDIA A40 GPU.

\subsection{Inference Procedure}

The fine-tuned model is applied to the remaining 1{,}000 
frames using greedy decoding with a maximum of 512 new tokens. 
Each inference call receives the same six-image prompt as 
training. The model outputs a structured JSON response 
containing tongue quality ratings, positioning judgements, 
best tongue, and best overall selections for all four variants. 
Outputs are parsed directly from the JSON response; malformed 
outputs are flagged and excluded from aggregation.

\section{Qualitative Video Inference Results}
\label{supp:video}

\begin{figure*}[h]
  \centering
  \includegraphics[width=\linewidth]{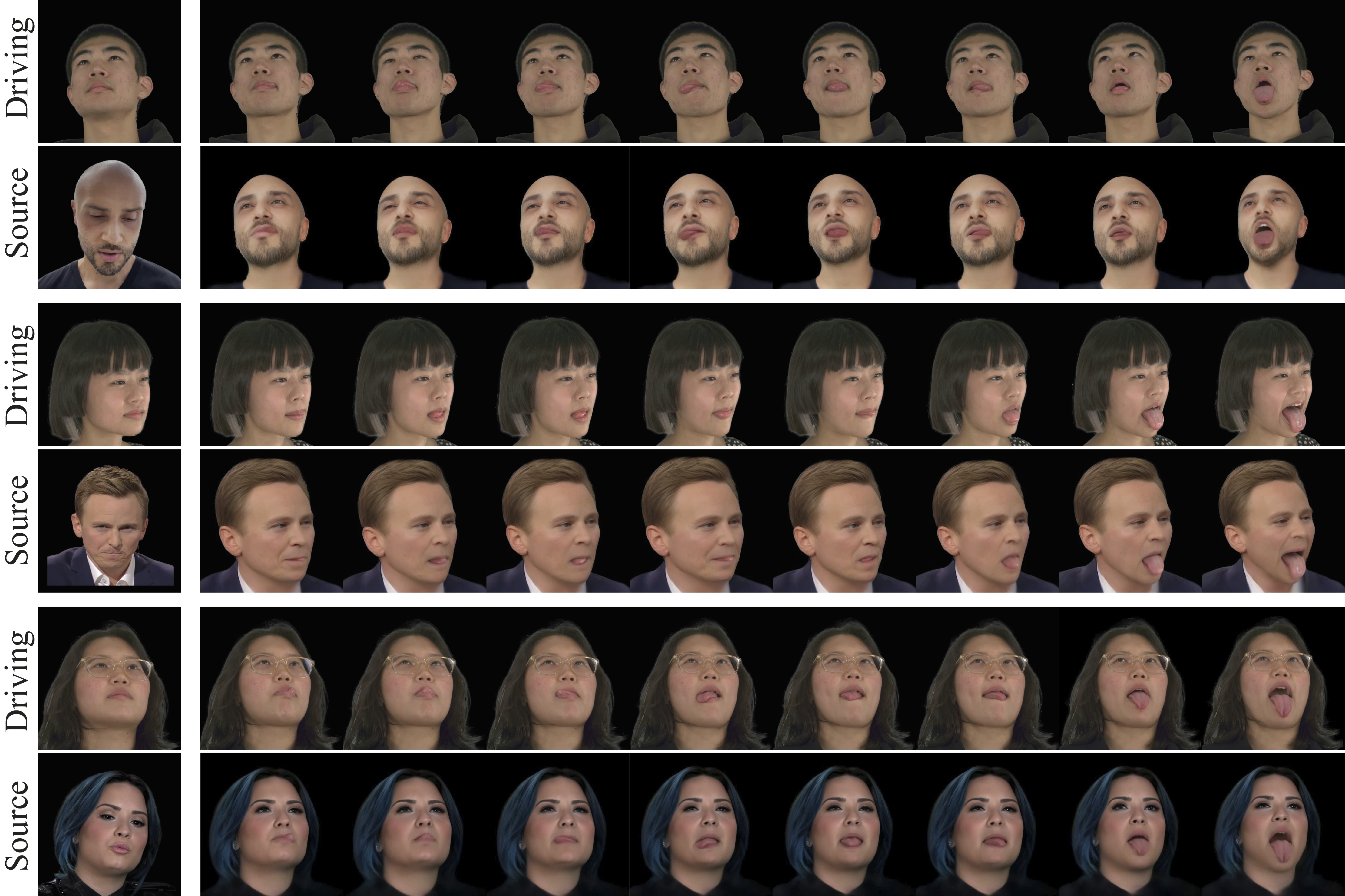}
  \caption{Consecutive frame strips from three source-driving 
  pairs. For each pair, the driving sequence (top) and our 
  generated output on the source identity (bottom) are shown 
  across eight frames. Our method produces temporally smooth 
  tongue dynamics that faithfully follow the driving identity's 
  articulation while preserving the source identity across all 
  non-synthesized facial regions.}
  \label{fig:supp_video}
\end{figure*}

Figure~\ref{fig:supp_video} shows consecutive frame strips 
from three source-driving pairs. Our method transfers tongue 
dynamics smoothly across frames while preserving the source 
identity's facial structure and head pose throughout each 
sequence.

%% file: main.bib
@String(ECCV  = {Eur. Conf. Comput. Vis.})

@String(ICLR  = {Int. Conf. Learn. Represent.})

@String(TOG   = {ACM Trans. Graph.})

@String(ECCV  = {ECCV})

@String(ICLR  = {ICLR})

@String(TOG   = {ACM TOG})

@article{kerbl20233d,
  title={3d gaussian splatting for real-time radiance field rendering.},
  author={Kerbl, Bernhard and Kopanas, Georgios and Leimk{\"u}hler, Thomas and Drettakis, George and others},
  journal={ACM Trans. Graph.},
  volume={42},
  number={4},
  pages={139--1},
  year={2023}
}

@inproceedings{dhamo2024headgas,
  title={Headgas: Real-time animatable head avatars via 3d gaussian splatting},
  author={Dhamo, Helisa and Nie, Yinyu and Moreau, Arthur and Song, Jifei and Shaw, Richard and Zhou, Yiren and P{\'e}rez-Pellitero, Eduardo},
  booktitle={European Conference on Computer Vision},
  pages={459--476},
  year={2024},
  organization={Springer}
}

@article{guo2024liveportrait,
  title={Liveportrait: Efficient portrait animation with stitching and retargeting control},
  author={Guo, Jianzhu and Zhang, Dingyun and Liu, Xiaoqiang and Zhong, Zhizhou and Zhang, Yuan and Wan, Pengfei and Zhang, Di},
  journal={arXiv preprint arXiv:2407.03168},
  year={2024}
}

@inproceedings{deng2024portrait4d,
  title={Portrait4d: Learning one-shot 4d head avatar synthesis using synthetic data},
  author={Deng, Yu and Wang, Duomin and Ren, Xiaohang and Chen, Xingyu and Wang, Baoyuan},
  booktitle={Proceedings of the IEEE/CVF Conference on Computer Vision and Pattern Recognition},
  pages={7119--7130},
  year={2024}
}

@article{chu2024gpavatar,
  title={GPAvatar: Generalizable and precise head avatar from image (s)},
  author={Chu, Xuangeng and Li, Yu and Zeng, Ailing and Yang, Tianyu and Lin, Lijian and Liu, Yunfei and Harada, Tatsuya},
  journal={arXiv preprint arXiv:2401.10215},
  year={2024}
}

@inproceedings{qian2024gaussianavatars,
  title={Gaussianavatars: Photorealistic head avatars with rigged 3d gaussians},
  author={Qian, Shenhan and Kirschstein, Tobias and Schoneveld, Liam and Davoli, Davide and Giebenhain, Simon and Nie{\ss}ner, Matthias},
  booktitle={Proceedings of the IEEE/CVF Conference on Computer Vision and Pattern Recognition},
  pages={20299--20309},
  year={2024}
}

@article{zhao2025x,
  title={X-nemo: Expressive neural motion reenactment via disentangled latent attention},
  author={Zhao, Xiaochen and Xu, Hongyi and Song, Guoxian and Xie, You and Zhang, Chenxu and Li, Xiu and Luo, Linjie and Suo, Jinli and Liu, Yebin},
  journal={arXiv preprint arXiv:2507.23143},
  year={2025}
}

@incollection{blanz2023morphable,
  title={A morphable model for the synthesis of 3D faces},
  author={Blanz, Volker and Vetter, Thomas},
  booktitle={Seminal Graphics Papers: Pushing the Boundaries, Volume 2},
  pages={157--164},
  year={2023}
}

@article{li2017learning,
  title={Learning a model of facial shape and expression from 4D scans.},
  author={Li, Tianye and Bolkart, Timo and Black, Michael J and Li, Hao and Romero, Javier},
  journal={ACM Trans. Graph.},
  volume={36},
  number={6},
  pages={194--1},
  year={2017}
}

@inproceedings{rombach2022high,
  title={High-resolution image synthesis with latent diffusion models},
  author={Rombach, Robin and Blattmann, Andreas and Lorenz, Dominik and Esser, Patrick and Ommer, Bj{\"o}rn},
  booktitle={Proceedings of the IEEE/CVF conference on computer vision and pattern recognition},
  pages={10684--10695},
  year={2022}
}

@article{chu2024generalizable,
  title={Generalizable and animatable gaussian head avatar},
  author={Chu, Xuangeng and Harada, Tatsuya},
  journal={Advances in Neural Information Processing Systems},
  volume={37},
  pages={57642--57670},
  year={2024}
}

@inproceedings{lugmayr2022repaint,
  title={Repaint: Inpainting using denoising diffusion probabilistic models},
  author={Lugmayr, Andreas and Danelljan, Martin and Romero, Andres and Yu, Fisher and Timofte, Radu and Van Gool, Luc},
  booktitle={Proceedings of the IEEE/CVF conference on computer vision and pattern recognition},
  pages={11461--11471},
  year={2022}
}

@article{kirschstein2023nersemble,
  title={Nersemble: Multi-view radiance field reconstruction of human heads},
  author={Kirschstein, Tobias and Qian, Shenhan and Giebenhain, Simon and Walter, Tim and Nie{\ss}ner, Matthias},
  journal={ACM Transactions on Graphics (TOG)},
  volume={42},
  number={4},
  pages={1--14},
  year={2023},
  publisher={ACM New York, NY, USA}
}

@inproceedings{kirillov2023segment,
  title={Segment anything},
  author={Kirillov, Alexander and Mintun, Eric and Ravi, Nikhila and Mao, Hanzi and Rolland, Chloe and Gustafson, Laura and Xiao, Tete and Whitehead, Spencer and Berg, Alexander C and Lo, Wan-Yen and others},
  booktitle={Proceedings of the IEEE/CVF international conference on computer vision},
  pages={4015--4026},
  year={2023}
}

@inproceedings{yu2018bisenet,
  title={Bisenet: Bilateral segmentation network for real-time semantic segmentation},
  author={Yu, Changqian and Wang, Jingbo and Peng, Chao and Gao, Changxin and Yu, Gang and Sang, Nong},
  booktitle={Proceedings of the European conference on computer vision (ECCV)},
  pages={325--341},
  year={2018}
}

@inproceedings{cao2023tonguesam,
  title={TongueSAM: An universal tongue segmentation model based on SAM with zero-shot},
  author={Cao, Shan and Wu, Qingfeng and Ma, Linjian},
  booktitle={2023 IEEE international conference on bioinformatics and biomedicine (BIBM)},
  pages={4520--4526},
  year={2023},
  organization={IEEE}
}

@inproceedings{lee2020maskgan,
  title={Maskgan: Towards diverse and interactive facial image manipulation},
  author={Lee, Cheng-Han and Liu, Ziwei and Wu, Lingyun and Luo, Ping},
  booktitle={Proceedings of the IEEE/CVF conference on computer vision and pattern recognition},
  pages={5549--5558},
  year={2020}
}

@inproceedings{xie2022vfhq,
  title={Vfhq: A high-quality dataset and benchmark for video face super-resolution},
  author={Xie, Liangbin and Wang, Xintao and Zhang, Honglun and Dong, Chao and Shan, Ying},
  booktitle={Proceedings of the IEEE/CVF Conference on Computer Vision and Pattern Recognition},
  pages={657--666},
  year={2022}
}

@inproceedings{shrivastava2016training,
  title={Training region-based object detectors with online hard example mining},
  author={Shrivastava, Abhinav and Gupta, Abhinav and Girshick, Ross},
  booktitle={Proceedings of the IEEE conference on computer vision and pattern recognition},
  pages={761--769},
  year={2016}
}

@article{song2020denoising,
  title={Denoising diffusion implicit models},
  author={Song, Jiaming and Meng, Chenlin and Ermon, Stefano},
  journal={arXiv preprint arXiv:2010.02502},
  year={2020}
}

@inproceedings{zhu2024champ,
  title={Champ: Controllable and consistent human image animation with 3d parametric guidance},
  author={Zhu, Shenhao and Chen, Junming Leo and Dai, Zuozhuo and Dong, Zilong and Xu, Yinghui and Cao, Xun and Yao, Yao and Zhu, Hao and Zhu, Siyu},
  booktitle={European Conference on Computer Vision},
  pages={145--162},
  year={2024},
  organization={Springer}
}

@inproceedings{radford2021learning,
  title={Learning transferable visual models from natural language supervision},
  author={Radford, Alec and Kim, Jong Wook and Hallacy, Chris and Ramesh, Aditya and Goh, Gabriel and Agarwal, Sandhini and Sastry, Girish and Askell, Amanda and Mishkin, Pamela and Clark, Jack and others},
  booktitle={International conference on machine learning},
  pages={8748--8763},
  year={2021},
  organization={PmLR}
}

@inproceedings{zhou2024headstudio,
  title={Headstudio: Text to animatable head avatars with 3d gaussian splatting},
  author={Zhou, Zhenglin and Ma, Fan and Fan, Hehe and Yang, Zongxin and Yang, Yi},
  booktitle={European Conference on Computer Vision},
  pages={145--163},
  year={2024},
  organization={Springer}
}

@inproceedings{ju2024brushnet,
  title={Brushnet: A plug-and-play image inpainting model with decomposed dual-branch diffusion},
  author={Ju, Xuan and Liu, Xian and Wang, Xintao and Bian, Yuxuan and Shan, Ying and Xu, Qiang},
  booktitle={European Conference on Computer Vision},
  pages={150--168},
  year={2024},
  organization={Springer}
}

@inproceedings{lin2024common,
  title={Common diffusion noise schedules and sample steps are flawed},
  author={Lin, Shanchuan and Liu, Bingchen and Li, Jiashi and Yang, Xiao},
  booktitle={Proceedings of the IEEE/CVF winter conference on applications of computer vision},
  pages={5404--5411},
  year={2024}
}

@inproceedings{deng2019arcface,
  title={Arcface: Additive angular margin loss for deep face recognition},
  author={Deng, Jiankang and Guo, Jia and Xue, Niannan and Zafeiriou, Stefanos},
  booktitle={Proceedings of the IEEE/CVF conference on computer vision and pattern recognition},
  pages={4690--4699},
  year={2019}
}

@inproceedings{xie2024x,
  title={X-portrait: Expressive portrait animation with hierarchical motion attention},
  author={Xie, You and Xu, Hongyi and Song, Guoxian and Wang, Chao and Shi, Yichun and Luo, Linjie},
  booktitle={ACM SIGGRAPH 2024 conference papers},
  pages={1--11},
  year={2024}
}

@inproceedings{hu2024animate,
  title={Animate anyone: Consistent and controllable image-to-video synthesis for character animation},
  author={Hu, Li},
  booktitle={Proceedings of the IEEE/CVF conference on computer vision and pattern recognition},
  pages={8153--8163},
  year={2024}
}

@inproceedings{zhang2018unreasonable,
  title={The unreasonable effectiveness of deep features as a perceptual metric},
  author={Zhang, Richard and Isola, Phillip and Efros, Alexei A and Shechtman, Eli and Wang, Oliver},
  booktitle={Proceedings of the IEEE conference on computer vision and pattern recognition},
  pages={586--595},
  year={2018}
}

@article{yang2025qwen3,
  title={Qwen3 technical report},
  author={Yang, An and Li, Anfeng and Yang, Baosong and Zhang, Beichen and Hui, Binyuan and Zheng, Bo and Yu, Bowen and Gao, Chang and Huang, Chengen and Lv, Chenxu and others},
  journal={arXiv preprint arXiv:2505.09388},
  year={2025}
}

@article{hu2022lora,
  title={Lora: Low-rank adaptation of large language models.},
  author={Hu, Edward J and Shen, Yelong and Wallis, Phillip and Allen-Zhu, Zeyuan and Li, Yuanzhi and Wang, Shean and Wang, Liang and Chen, Weizhu and others},
  journal={Iclr},
  volume={1},
  number={2},
  pages={3},
  year={2022}
}
